\pdfoutput=1
\documentclass[11pt]{article}

\usepackage[final]{acl}

\usepackage{times}
\usepackage{latexsym}

\usepackage[T1]{fontenc}
\usepackage[utf8]{inputenc}

\usepackage{microtype}

\usepackage{inconsolata}

\usepackage{graphicx}
\usepackage{tabularx}
\usepackage{array}
\usepackage{subcaption}
\usepackage{pgfplots}
\usepackage{afterpage}
\usepgfplotslibrary{groupplots}
\usepgfplotslibrary{colorbrewer}
\usepackage{tikz}
\usepackage{booktabs}
\usepackage{multirow}
\usepackage{xcolor}
\usetikzlibrary{matrix}
\usepackage{hyperref}

\title{Detecting Machine-Generated Texts: Not Just ``AI vs Humans'' and Explainability is Complicated}

\author{Jiazhou Ji$^{1\ast}$ \quad Ruizhe Li$^{2\ast}$\quad Shujun Li$^{3}\thanks{Equal contributions.}$ \quad Jie Guo$^{1}$\thanks{Corresponding author: \texttt{guojie@sjtu.edu.cn}}\\
\textbf{Weidong Qiu$^{1}$\quad Zheng Huang$^{1}$\quad Chiyu Chen$^{1}$\quad Xiaoyu Jiang$^{1}$ \quad Xinru Lu$^{1}$} \\
$^1$School of Cyber Science and Engineering, Shanghai Jiao Tong University, China\\ $^2$Department of Computing Science, University of Aberdeen, UK\\ $^3$Institute of Cyber Security for Society (iCSS) \& School of Computing, University of Kent, UK\\
}

\begin{document}

\maketitle

\begin{abstract}
As Large Language Models (LLMs) rapidly advance, increasing concerns arise regarding risks about the actual authorship of texts we see online and in the real world. The task of distinguishing LLM-authored texts is complicated by the nuanced and overlapping behaviors of both machines and humans. In this paper, we challenge the current practice of considering the LLM-generated text detection a binary classification task of differentiating human from AI. Instead, we introduce a novel ternary text classification scheme, adding an ``undecided'' category for texts that could be attributed to either source, and we show that this new category is crucial to understand how to make the detection result more explainable to lay users. This research shifts the paradigm from merely classifying to explaining machine-generated texts, emphasizing the need for detectors to provide clear and understandable explanations to users. Our study involves creating four new datasets comprised of texts from various LLMs and human authors. Based on the new datasets, we performed binary classification tests to ascertain the most effective state-of-the-art (SOTA) detection methods and identified SOTA LLMs capable of producing harder-to-detect texts. Then, we constructed a new dataset of texts generated by the two top-performing LLMs and human authors, and asked four human annotators to produce ternary labels with explanation notes. This dataset was used to investigate how three top-performing SOTA detectors behave in the new ternary classification context. Our results highlight why the ``undecided'' category is much needed from the viewpoint of explainability. Additionally, we conducted an analysis of explainability of the three best-performing detectors and the explanation notes of the human annotators, revealing insights about the complexity of explainable detection of machine-generated texts. Finally, we propose guidelines for developing future detection systems with improved explanatory power.

% \noindent\textbf{Key words: }machine-generated text detection, ternary classification, explainability
\end{abstract}

\section{Introduction}

With the rapid evolution of Large Language Models (LLMs) such as ChatGPT-4~\cite{OpenAI2023GTP4}, the sophistication and human-like quality of texts generated by these models have notably increased, enabling them to produce diverse content in response to specific prompts. These advancements bring not only numerous practical applications but also raise significant challenges including potential academic fraud and actual authorship. Extensive research has been undertaken to differentiate between machine-generated texts (MGTs) and human-generated texts (HGTs), primarily employing model-based approaches~\cite{Wang2023SeqXGPT, Bhattacharjee2023ConDA, Rezaei2024CLULab} and statistical methods that analyze inherent text characteristics~\cite{Hans2024Binoculars, Bao2024Fast-DetectGPT, zhang2024detecting, Ma2024TOCSIN}. Several online platforms such as GPTZero~\cite{gptzero_website} and Sapling~\cite{sapling_website} have also demonstrated robust capabilities in differentiating MGTs from HGTs.

Traditionally, the detection of MGTs has relied on a binary classification framework that discerns between MGTs and HGTs. However, the boundaries between MGTs and HGTs are increasingly ambiguous due to the rapid enhancements in LLMs, thereby complicating the effectiveness of simple binary classification systems. For instance, in statistical detection, the characteristics of a given MGT might deviate significantly from typical MGTs patterns and mirror those of HGTs, leading to a misclassification. Model-based methods often struggle with generalization as they tend to learn features that are specific to the data they are trained on (usually limited to one or more specific LLMs), which may not necessarily work as new models emerge. Moreover, many existing detection systems lack an explainability component. Recent studies have emphasized the importance of interpretability, introducing methods like LIME-based explanations~\cite{Joshi2024HULLMI} and SHAP~\cite{mitrovic2023chatgpt} to enhance user understanding. However, their effectiveness in providing meaningful insights appears limited according to our evaluations of one such methods (an online closed-source detector) GPTZero~\cite{gptzero_website}. This shortfall emphasizes a critical gap: the need for enhanced explainability in MGT detectors to improve end users' trust in such systems.

In order to address these limitations, our study introduces a novel ternary classification system for analyzing texts. Recognizing that some texts may simultaneously share characteristics of both MGTs and HGTs, we have added an ``undecided'' category to our classification framework. This category accounts for several complex cases, including stylistically blended texts co-authored by humans and LLMs, inherently ambiguous writing that plausibly fits either class, and borderline cases where weak signals exist but lack robustness. To support this classification, we developed a ternary dataset and designed experiments to test the validity of this approach. Our methodology not only includes rigorous statistical and model-based analyses, but also incorporates detailed human evaluations to provide a nuanced understanding of the new ternary text classification task and the complexity of producing human-understandable explanations. By comparing the explanatory power of human assessments with that of automated detectors, we highlight the current explanatory limitations faced by MGT detectors.

Through some binary classification experiments based on four new datasets covering multiple state-of-the-art (SOTA) LLMs, we established that the most advanced LLMs currently available are ChatGPT-4 and ChatGPT-3.5, in terms of defeating multiple SOTA MGT detectors. The detectors that performed the best in our experiments are GPTZero~\cite{gptzero_website}, Sapling~\cite{sapling_website} and Binoculars~\cite{Hans2024Binoculars}. Building on these findings, we crafted a ternary classification dataset using texts from the aforementioned top-performing LLMs. We organized human coders to annotate these texts, applying the ternary classification framework and providing detailed explanations for their decisions. Subsequent experiments with the top three detectors proved the limitations of binary classification so that the new ``undecided'' category should be seriously considered in future research on MGT detection. Our comparative analysis between the human-provided explanations and those offered by the detector GPTZero illuminated significant gaps in current automated explanations. While human explanations provide valuable insights, they also exhibit inherent limitations and imply the complexity and challenges behind developing more explainable MGT detectors.

In conclusion, our research not only challenges existing paradigms in MGT detection but also sets a foundation for future innovations in detector design, particularly in enhancing explainability. This work suggests new directions for the development of detection systems that are not only effective but also transparent and interpretable to users.

% In this work, our objective is to explore the current issues with text detectors and to propose novel ideas for improvement. In Section 2, we analyze the existing research on text detectors, focusing on their explainability and the current state of research related to this aspect. In Section 3, we design a binary classification experiment to test the performance of current detectors and the text generation capabilities of various LLMs. In Section 4, we organize a dataset using a human-generated three-category classification and conduct a ternary classification test on the detectors selected in Section 3. In Section 5, we examine the explainability of current detectors, the explanations provided by human coders, and identify ongoing issues. In Section 6, we discuss the results and limitations of our work and offer some thoughts and recommendations for future research.

\section{Related Work}

\subsection{Open-Source Detectors}

\paragraph{Zero-shot detection.} This approach leverages unique statistical properties distinguishing MGTs from HGTs. Past studies have employed various linguistic model-derived characteristics, such as entropy~\cite{He2023MGTBench}, average log-probability scores~\cite{solaiman2019release}, perplexity~\cite{Wu2023LLMDet}, and token cohesiveness~\cite{Ma2024TOCSIN} as useful statistical properties for detection. With the evolution of LLMs that generate increasingly sophisticated texts, more recent zero-shot detection strategies~\cite{gehrmann2019gltr, Mitchell2023DetectGPT, su2023detectllm, wu2023mfd, Bao2024Fast-DetectGPT, Kumari2024DEMASQ} have adapted to discern high-order features of advanced text generators. Notably, the Binoculars model~\cite{Hans2024Binoculars} leverages LLMs to perform next-token predictions at each text position, utilizing the log perplexity ratio compared to the baseline text as a distinguishing statistic.

\paragraph{Model-based detection.} This approach involves adapting existing models to learn from specific datasets for MGT detection~\cite{openai_text_classifier, He2023MGTBench}. It often includes sentence-level detection and analyses different LLM outputs~\cite{Wang2023SeqXGPT, Bhattacharjee2023ConDA, Antoun2024text2source, Rezaei2024CLULab}. However, these methods can suffer from overfitting and generally exhibit limited effectiveness in detecting texts across various domains.
%openai_text_classifier should appear here.

\paragraph{Other approaches.} There are also other approaches based on watermarking, adversarial learning based training, and human assistance~\cite{Wu2024survey}. These approaches are more complicated and are often a mixture of different approaches, so in this paper we consider two basic approaches only to make our work more focused.

\subsection{Online Close-Source Detection Systems}

Despite their closed-source nature, online detectors are of significant interest in academic research~\cite{Yang2023a}. For instance, GPTZero~\cite{gptzero_website} integrates several analytical components that predict if a piece of text is generated by machine or human with a confidence score, together with a sentence-by-sentence analysis capability. Similarly, Sapling~\cite{sapling_website} utilizes a transformer-based architecture akin to those found in generative AI systems. Moreover, various platforms offer an online MGT detection tool for all to use~\cite{originality_ai_website, copyleaks_website, crossplag_website, zerogpt_website}.

\subsection{Explainability in Current Detectors}

Recent studies have introduced methods to enhance the interpretability of detection models. For instance, the HULLMI framework employs LIME to provide insights into model predictions~\cite{Joshi2024HULLMI}. According to its official documentation~\cite{gptzero_website}, GPTZero uses the following six features to achieve explainability: readability, percent SAT, simplicity, perplexity, burstiness, and average sentence length. However, it does not provide clarity on how these features influence its final judgments. Other efforts have focused on integrating explanatory modules into detectors. One study~\cite{mitrovic2023chatgpt} implemented Shapley Additive Explanations (SHAP)~\cite{lundberg2017unified}, which assigns importance values to each feature, enhancing the interpretability of decisions in text source detection. Another investigation~\cite{andre2023detecting} computed textual attributes such as perplexity, grammar, and n-gram distributions to measure their effects on detection outcomes. Despite these advancements, the current state of detector explainability remains challenging for lay users to comprehend.

\section{Binary Classification Evaluation of Detectors on MGTs and HGTs}
\label{sec:binary_performance}

This section outlines the assessment of state-of-the-art (SOTA) MGT detectors through binary classification tests on datasets containing both MGTs and HGTs. Our objective is to identify the most effective and consistently accurate detectors across various datasets and to pinpoint LLMs that exhibit the strongest generative abilities and human-like output. This process will involve binary classification trials using custom-built datasets. The selected detectors and LLMs will then be utilized in further experimental investigations.

\subsection{Experimental Design}

We conducted our experiments using four datasets specifically constructed for this study. It is crucial to carefully select LLMs for text generation and appropriate sources of HGTs to assemble the dataset. We chose a mix of open-source and closed-source SOTA MGT detectors for evaluation and used standard performance metrics for the binary classification tests.

\paragraph{LLMs.} For text generation, we have opted for widely recognized models including the closed-source GPT-4o \cite{OpenAI2024GTP4O}, known for its robust performance. Additionally, we selected Google's Gemini Pro~\cite{gemini2023}, renowned for its ability to produce coherent and high-quality natural language outputs. From the open-source domain, we have chosen LLaMA3.3-70B \cite{dubey2024llama}, and Qwen2-72B \cite{yang2024qwen2}

\paragraph{HGT sources.} To ensure a diverse and representative collection of HGTs, we included selections from public datasets such as the HC3 dataset~\cite{Guo2023ChatGPT_vs_humans}, which contains texts from four other public Q\&A datasets and data crawled from Wikipedia. Notably, it includes a category of texts, similar to the ELI5 (``Explain Like I'm Five'') format~\cite{Fan2019ELI5}, where complex issues are explained in simple terms. We also extracted short texts from the IDMGSP dataset~\cite{abdalla2023benchmark}, which comprises titles, abstracts, introductions and conclusions of human-authored scientific papers, alongside texts manually selected from X/Twitter using tag searches to cover topics of everyday discourse. This blend of sources provides a broad spectrum of topics and writing styles in the human-generated texts within our dataset.
\begin{table*}[tb]
\centering
\small
\begin{tabularx}{\textwidth}{@{}c *{8}{>{\centering\arraybackslash}X} @{}}
\toprule
\multirow{2}{*}{Models} & \multirow{2}{*}{Accuracy} & \multicolumn{3}{c}{Machine as Positive} & \multicolumn{3}{c}{Human as Positive} & \multirow{2}{*}{Macro F1} \\
\cmidrule(lr{0.5em}){3-5} \cmidrule(lr{0.5em}){6-8}
& & Precision & Recall & F1 & Precision & Recall & F1 & \\
\midrule
GPTZero & 93.90\% & 92.87\% & 95.10\% & 93.97\% & 94.98\% & 92.70\% & 93.83\% & 93.90\% \\
Sapling & 92.65\% & 89.82\% & 96.20\% & 92.90\% & 95.91\% & 89.10\% & 92.38\% & 92.64\% \\
Binoculars & 86.75\% & 80.35\% & 97.30\% & 88.01\% & 96.58\% & 76.20\% & 85.19\% & 86.60\% \\
Fast-DetectGPT & 76.45\% & 86.38\% & 62.80\% & 72.73\% & 70.78\% & 90.10\% & 79.28\% & 76.00\% \\
MMD-MP & 74.05\% & 95.81\% & 50.30\% & 65.97\% & 66.31\% & 97.80\% & 79.03\% & 72.50\% \\
DEMASQ & 61.75\% & 57.60\% & 89.10\% & 69.96\% & 75.94\% & 34.40\% & 47.35\% & 58.66\% \\
DetectGPT & 54.80\% & 81.58\% & 12.40\% & 21.53\% & 52.60\% & 97.20\% & 68.26\% & 44.89\% \\
\bottomrule
\end{tabularx}
\caption{Binary classification performance of different detectors on the dataset of GPT-4o}
\label{tab:binary_performance_GPT4o}
\end{table*}

\begin{figure*}[tb]  % 't' 选项尝试将图表放置于页面顶部
\centering
\begin{tikzpicture}
\begin{axis}[
    ybar,
    width=\linewidth, % 使用整个文本宽度
    bar width=8pt, % 控制条形宽度
    height=0.35\linewidth, % 调整高度
    enlargelimits=0.15, % 扩大边界，给图表和轴标签更多空间
    legend style={at={(0.5,0.98)}, % 改变位置，将图例移至内部上方但在边框内
      anchor=north,legend columns=-1, font=\small, draw=none}, % 调整图例位置，避免与条形图重叠
    ylabel={Macro F1 Score (\%)}, % y轴标签
    ytick={0, 20, 40, 60, 80, 100},
    symbolic x coords={ChatGPT-4 (D1), Gemini Pro (D2), LLaMA3.3-70B (D3), Qwen2-72B (D4)}, % x轴坐标
    tick label style={font=\small},
    label style={font=\small},
    xtick=data,
    nodes near coords align={vertical}, % 数值标签垂直对齐
    legend image code/.code={
        \draw[draw=none] (0cm,-0.1cm) rectangle (0.3cm,0.1cm); % 取消图例中小条形图的边框
    },
    ]
\addplot+[ybar, fill=yellow!50, draw=none] coordinates {(ChatGPT-4 (D1),93.97) (Gemini Pro (D2),95.06) (LLaMA3.3-70B (D3),93.18) (Qwen2-72B (D4),92.00) };
\addplot+[ybar, fill=cyan!50, draw=none] coordinates {(ChatGPT-4 (D1),92.90) (Gemini Pro (D2),93.47) (LLaMA3.3-70B (D3),93.88) (Qwen2-72B (D4),93.16) };
\addplot+[ybar, fill=blue!50, draw=none] coordinates {(ChatGPT-4 (D1),88.01) (Gemini Pro (D2),88.07) (LLaMA3.3-70B (D3),88.82) (Qwen2-72B (D4),86.94) };
\addplot+[ybar, fill=orange!50, draw=none] coordinates {(ChatGPT-4 (D1),72.73) (Gemini Pro (D2),84.44) (LLaMA3.3-70B (D3),82.02) (Qwen2-72B (D4),74.97) };
\addplot+[ybar, fill=purple!50, draw=none] coordinates {(ChatGPT-4 (D1),65.97) (Gemini Pro (D2),77.53) (LLaMA3.3-70B (D3),80.82) (Qwen2-72B (D4),69.36) };
\addplot+[ybar, fill=red!50, draw=none] coordinates {(ChatGPT-4 (D1),69.96) (Gemini Pro (D2),68.99) (LLaMA3.3-70B (D3),70.93) (Qwen2-72B (D4),70.73) };
\addplot+[ybar, fill=green!50, draw=none] coordinates {(ChatGPT-4 (D1),21.53) (Gemini Pro (D2),56.92) (LLaMA3.3-70B (D3),49.60) (Qwen2-72B (D4),23.22) };
\legend{GPTZero, Sapling, Binoculars, Fast-DetectGPT, MMD-MP, DEMASQ, DetectGPT} % 图例标签
\end{axis}
\end{tikzpicture}
\caption{Comparison of detector performance across the four datasets produced by various LLMs, with MGTs as positive samples. The x-axis represents different datasets, while different bars represent different detectors.} % 图表标题
\label{fig:binary_detectors_performance}
\end{figure*}
\paragraph{MGT Detectors.} Initially, we chose GPTZero and Sapling as the leading commercial (online and closed-source) detectors from the proprietary sector. We then extended our selection to include several notable open-source detectors such as Binoculars~\cite{Hans2024Binoculars}, Fast-DetectGPT~\cite{Bao2024Fast-DetectGPT}, MMD-MP~\cite{zhang2024detecting}, DEMASQ~\cite{Kumari2024DEMASQ}, and DetectGPT~\cite{Mitchell2023DetectGPT}.

\paragraph{Custom-built Datasets.} Four datasets were built using the selected LLMs and HGT sources, as detailed in Table~\ref{tab:binary_datasets}. To control variables in subsequent analyses, the HGTs within these datasets were maintained consistently across all experiments. This standardization can help isolate the variable effects of different LLM outputs on detector performance.

\paragraph{Evaluation metrics.} The detectors are expected to maximize MGT detection accuracy while minimizing false positives among HGTs. Therefore, Precision, Recall, and F1 scores for MGTs are selected as primary evaluation metrics. Other metrics, such as the macro F1 score across two classification situations (MGTs and HGTs as positive samples, respectively), have also been used to provide a comprehensive assessment of detector performance.

\begin{table}[tb]
\centering
\small
\begin{tabular}{ccccccc}
\toprule
Dataset & MGTs & HGTs\\
\midrule
D1 & 1000 (generated by GPT-4o) & 1000\\
D2 & 1000 (generated by Gemini Pro) & 1000\\
D3 & 1000 (generated by LLaMA3.3-70B) & 1000\\
D4 & 1000 (generated by Qwen2-72B) & 1000\\
\bottomrule
\end{tabular}
\caption{Composition of the four datasets. The texts cover a wide range of topics including economics, healthcare, science, literature, sports, and daily life.}
\label{tab:binary_datasets}
\end{table}

\subsection{Results}

We evaluated various detectors on datasets, as detailed in Table~\ref{tab:binary_performance_GPT4o}, focusing on the dataset generated by GPT-4o. This table highlights the performance of detectors using both humans and machines as the positive label. The results indicate that online detectors, GPTZero and Sapling, significantly outperform local open-source counterparts. Specifically, DEMASQ effectively identifies MGTs but struggles with HGT detection. Conversely, DetectGPT shows limited capability in detecting MGTs while performing adequately with HGTs. See Appendix~\ref{appendix:binary_performance_results} for extended results for other datasets.

Figure~\ref{fig:binary_detectors_performance} visually compares F1 scores of all tested MGT detectors across all four datasets, confirming the superior performance of GPTZero and Sapling over local models. Among the latter, Binoculars ranks the highest, demonstrating a consistent performance across all datasets, suggesting its being less susceptible to overfitting compared to other local models. Further analysis reveals that texts generated by GPT-4o and Qwen2-72B are generally more challenging to classify across all detectors, compared to those generated by Gemini Pro and LLaMA3.3-70B, implying that GPT-4o and Qwen2-72B can produce texts that more closely resemble human writing. Based on these findings, for the further experiments and discussions about the new ternary classification framework and the complexity of explainability, we chose to use a mixed dataset with texts generated by GPT-4o and Qwen2-72B, and HGTs. Similarly, on the selection of MGT detectors, we focused on three top-performing ones, GPTZero, Sapling, and Binoculars.

\section{Ternary Classification Tests for Selected MGT Detectors}

\subsection{Manual Annotation and Explanations}

Following the outcomes from binary classification experiments, we compiled a new dataset containing texts from GPT-4o, Qwen2-72B and human authors. The dataset consists of 400 texts, with 100 from GPT-4o, 100 from Qwen2-72B, and 200 from human authors. We recruited four individuals from the English-speaking community to annotate 400 texts to categorize each text into one of three groups: human, machine, and undecided. The occupations of the four English-native speakers are a senior master student of computer science, a lecturer of English Language and Literature at a UK University, a senior PhD student of NLP and a lecturer of Law at a UK University, respectively, which indicates the great diversity and qualifications of those human annotators. They also provided explanatory notes to justify their annotation results. Each annotator first independently annotated the whole 400 texts and provided their explanations. 

After all four annotators finished their work, we calculated Fleiss' kappa~\cite{Fleiss1971kappa}, which was 0.3760, indicating fair agreement among the annotators. To address the disagreements, all annotators entered into a collaborative discussion on the texts with different opinions, without revealing the ground truth to the annotators, and the annotators were asked to refine their annotations. After the annotations were updated, we calculated Fleiss' kappa again, which increased to 0.9875, reflecting a near-complete consensus among all annotators. Any texts that remain to have no consensus were labeled as ``undecided''. The explanation notes of the four annotators were merged and refined to be more consistent after the first author discussed with the four annotators and other co-authors. More details of the dataset can be found in Table~\ref{tab:comparison_human_and_gt}, which shows that all human annotated MGTs and HGTs are 96.80\% correct according to the ground truth labels. The high percentage of undecided texts itself is indicative and already shows that the traditional binary classification approaches may be problematic. More information about how the human annotators' work is given in Appendix~\ref{appendix:human_annotation}.

The human annotation results revealed that, although some automated MGT detectors have achieved very good performance in predicting ground truth labels, human annotators were clearly not convinced by the cases falling into the ``undecided'' category. This can be partly explained by what an ideal machine-based text generator is supposed to do -- to produce texts that are HGTs. Although we may argue that SOTA LLM-based generators are still far from ideal, the human annotators clearly have seen many example MGTs that are sufficiently human-like so that there is no convincing way to label them as just MGTs or HGTs, so ``undecided'' would be a better class to describe them.

\begin{table}[tb]
\centering
\small
\begin{tabular}{@{}cccc@{}}
\toprule
Human Annotation & Total & GT: Machine & GT: Human\\
\midrule
Machine & 162 & 159 & 3 \\
Human & 182 & 8 & 174 \\
Undecided & 56 & 33 (58.93\%) & 23 (41.07\%)\\
\bottomrule
\end{tabular}
\caption{Comparison between human annotations and ground truth (GT) labels.}
\label{tab:comparison_human_and_gt}
\end{table}

\subsection{Method}

Using the new dataset with ternary labels, we investigated how the three top-performing binary MGT detectors performed in the context of the ternary classification task. We generated $3\times 2$ confusion matrices to observe how the three different types of texts, particularly those in the new ``undecided'' category, are classified by the MGT detectors.
\pgfplotsset{
    CM common axis styles/.style={
        width=1.8in,
        height=1.4in,
        xmajorgrids=false,
        ymajorgrids=false,
        xmin=1, xmax=3,
        ymin=1, ymax=2,
        xtick={1,2,3},
        ytick={1,2},
        xticklabels={Machine, Human, Undecided},
        yticklabels={Machine, Human},
        x tick label style={font=\scriptsize},
        y tick label style={font=\scriptsize},
        tickwidth=0pt,
        tick align=outside,
        enlargelimits={abs=0.5},
        colormap={whiteblue}{rgb255(0cm)=(255,255,255) rgb255(1cm)=(128,128,255)}
    },
    CM common plot styles/.style={
        matrix plot,
        mesh/cols=3,
        mesh/rows=2,
        point meta=explicit,
        nodes near coords,
        nodes near coords align={center},
        visualization depends on={value \thisrow{q} \as \rawvalue},
        nodes near coords style={font=\tiny, color=black}
    }
}

\begin{figure*}[tb]
\begin{subfigure}{0.32\textwidth}
    \centering
    \begin{tikzpicture}
    \begin{axis}[
    CM common axis styles
    ]
    \addplot [
    CM common plot styles
    ] table [meta=p] {
    x y p
    1 1 95.7
    2 1 3.3
    3 1 69.6
    1 2 4.3
    2 2 96.7
    3 2 30.4
    };
    \end{axis}
    \end{tikzpicture}
    \caption{GPTZero}
\end{subfigure}%
\begin{subfigure}{0.32\textwidth}
    \centering
    \begin{tikzpicture}
    \begin{axis}[
    CM common axis styles
    ]
    \addplot [
    CM common plot styles
    ] table [meta=p] {
    x y p
    1 1 98.8
    2 1 15.9
    3 1 75.0
    1 2 1.2
    2 2 84.1
    3 2 25.0
    };
    \end{axis}
    \end{tikzpicture}
    \caption{Sapling}
\end{subfigure}%
\begin{subfigure}{0.32\textwidth}
    \centering
    \begin{tikzpicture}
    \begin{axis}[
    CM common axis styles,
    colorbar,
    colorbar style={
        width=2mm,
        ytick={0, 20, 40, 60, 80, 100},
        yticklabel style={font=\scriptsize},
        title=\%,
        title style={yshift=-2mm}
    },
    point meta min=0,
    point meta max=100
    ]
    \addplot [
    CM common plot styles
    ] table [meta=p] {
    x y p
    1 1 94.4
    2 1 21.4
    3 1 85.7
    1 2 5.6
    2 2 78.6
    3 2 14.3
    };
    \end{axis}
    \end{tikzpicture}
    \caption{Binoculars}
\end{subfigure}
\caption{
Confusion matrices of three binary detectors (GPTZero, Sapling, and Binoculars) applied to a ternary classification task with labels \textit{Machine}, \textit{Human}, and \textit{Undecided}. All input texts were sufficiently long to meet the minimum input length required by all detectors, so no predictions were skipped.
}
\label{fig:ternary_confusion_matrices}
\end{figure*}
\begin{table*}[tb]
\centering
\small
\begin{tabular}{@{}p{\linewidth}@{}} % 使用p{width}指定每列宽度
\toprule
\textbf{Source:} ChatGPT-4

\textbf{Text:} Sweating itself does not directly cause colds. Colds are caused by viruses, not by being cold or sweating. However, if you sweat and then get chilled, this might weaken your immune system temporarily, making you more susceptible to catching a cold virus. Additionally, the belief that sweating leads to colds might stem from confusing the symptoms of a cold, which can include sweating, with the cause of the cold.\\
\midrule
\textbf{GPTZero result}: AI

\textbf{GPTZero explanations}: Readability: 72.3 (Medium) | Percent SAT: 1.7 (Medium) | Simplicity: 35.2 (Medium) | Perplexity: 45.3 (Medium) | Burstiness: 37.9 (Medium) | Average sentence length: 22.3 (Medium)\\
\midrule
\textbf{Human labels}: undecided

\textbf{Human explanations}: The text is free from grammatical and spelling errors. This passage elucidates the relationship between sweating and colds, maintaining an objective and rigorous tone. It encompasses both common knowledge and scientific principles. The structure of the text is clear, with adverbial usage enhancing the clarity and fluency of the sentences. The text avoids unnecessary repetition, making it readily comprehensible. Therefore, it should be categorized as ``undecided.''\\
\bottomrule
\end{tabular}
\caption{Comparison between abstract scores from GPTZero and human-readable explanations}
\label{tab:example}
\end{table*}
\subsection{Results}

The confusion matrices for the detectors GPTZero, Sapling, and Binoculars, detailed in Fig.~\ref{fig:ternary_confusion_matrices}, reveal that, while the detection accuracy is high for clearly defined MGTs and HGTs (which was expected based on the results of the binary classification experiments reported in the previous section), challenges persist with the ``undecided'' texts.
%Notably, GPTZero misclassified only 3\% samples as MGTs or HGTs, showing improvement in its latest iteration with only 1\% error, as discussed in Appendix~\ref{sec:appendixf}. However, both Sapling and Binoculars occasionally mislabeled human-produced texts as machine-generated.
The most interesting pattern is that all three detectors are clearly biased on texts labeled as ``undecided'': they all have a clear tendency to classify such texts as MGTs. This bias is largely aligned with the biased percentage of MGTs in the ``undecided'' category as shown in Table~\ref{tab:comparison_human_and_gt}. Considering that human annotators considered such texts difficult to judge, it is likely also difficult for the MGT detectors to explain why they consider such texts generated by either machines or humans. Another interesting observation is that, both Sapling and Binoculars have a much higher error rate for HGTs than for MGTs labeled by our human annotators, implying HGTs may be generally harder to detect than MGTs for most detectors. GPTZero does not seem to suffer from this problem, but due to its closed-source nature it is unclear how it achieved such a performance.

\section{Explainability of Detectors}

%In previous sections, we introduced a ternary classification framework that includes an ``undecided'' category, and we assessed the effectiveness of leading binary classifiers using this model on a manually annotated dataset. The results revealed limitations in the persuasiveness of these experiments because when texts categorically identified as either machine-generated or human-generated are classified ``undecided'' by experts, while detectors make definitive classifications, we cannot conclusively argue that experts surpass detectors in recognizing and understanding the nature of the generated texts. The key to justifying the necessity of the “undecided” category and highlighting the shortcomings of existing detectors lies in the quality of the explanations they provide.

The results in the previous section indicate the importance for binary MGT detectors to explain their results to human users, which is particularly important for texts in the ``undecided'' category since human users may not agree on binary labels for such texts, not mentioning the results from an automated MGT detector. In this section, we report our analysis of explanation notes given by the four human annotators who constructed the ternary dataset we used.

\subsection{Analysis of GPTZero's Explainability}

%During our ternary classification tests, Sapling functioned as a black box, providing no explanatory feedback, thereby lacking in transparency. Similarly, while Binoculars is open-source, its code analysis shows that it simply outputs a decision based on a threshold, which also does not enhance its explainability.

% Different from Sapling and Binoculars, which do not provide any explanation to their results, GPTZero offers the following six concrete metrics to offer some level of explainability to their results: readability, percent SAT words, simplicity, perplexity, burstiness, and average sentence length. Other than giving values of the metrics, it does not clarify how they affect its decision-making process.

Different from Sapling and Binoculars, which do not provide any explanation for their results, earlier versions of GPTZero (as of March 2024) offered six concrete metrics to support a degree of explainability: readability, percent SAT words, simplicity, perplexity, burstiness, and average sentence length. However, these features have since been removed, and GPTZero now provides no explanation for its decisions, leaving users with binary or scalar outputs without interpretability.

% GPTZero's metric definitions, sourced from its documentation, include readability affected by word length and syllable count, percent SAT as the occurrence of college-level vocabulary, and simplicity as the use of common words. Perplexity and burstiness assess sentence predictability, while average sentence length provides a straightforward measure. 

Table~\ref{tab:example} shows an example, comparing the six explainability metrics used by GPTZero and the explanation notes given by our human annotators. As can be seen, the metrics used by GPTZero has limited explanatory power because they are too abstract. For instance, all the six metrics are marked as ``Medium'', which does not explain why the final judgment is AI. Instead, ``Medium'' may better fit into the ``undecoded'' category of our ternary classification framework, as what the human annotators stated in their more human-understandable explanation notes.

%On the contrary, human annotators offer a qualitative review discussing the text’s accuracy, clarity, structure, and style in layman’s terms, enhancing the user’s comprehension of how the text conveys its message. This approach proves more beneficial for general understanding than abstract statistical data.

A further empirical analysis was performed to study how the six explainability metrics claimed by GPTZero affect the final results. We constructed a dataset using texts in the datasets we used in previous sections, and used the six metrics as the input features and the GPTZero's detection results as the target class labels. We used an 80-20 training-testing split and applied various traditional machine learning models including logistic regression~\cite{Cox1958}, SVC~\cite{Cortes1995SVN}, perceptron~\cite{Rosenblatt1958perceptron}, and decision tree~\cite{Breiman1984}. The results  showed that two metrics, Readability and Perplexity, significantly affect the GPTZero's decision-making, while other metrics played a minor role. Yet, the accuracy rates of all models stayed below 80\%, implying that GPTZero uses other features and/or mechanisms to achieve its much higher performance observed in Section~\ref{sec:binary_performance}. For a comprehensive breakdown of these results, refer to Appendix~\ref{appendix:GPTZero_explainability}.

\subsection{Explanation Categories Provided by Human Annotators}

%Human experts, unlike most current detectors, base their text source determinations on explicit text features rather than abstract statistical metrics.
Our analysis of human annotators' explanation notes revealed six primary categories, each detailed in Appendix~\ref{appendix:human_explanation_categories}.

\paragraph{Linguistic fluency and coherence.} This refers to the correctness and naturalness of language, including grammar, phrasing, punctuation, and sentence variety. HGTs may include occasional errors or awkward expressions, reflecting human fallibility. In contrast, MGTs often produce technically correct but overly polished language, which may lack the irregularities typical of human writing.

\paragraph{Stylistic register and tone.} This refers to the emotional tone and formality of a text, including the use of personal pronouns, colloquial language, or conversational expressions. HGTs are more likely to display personal or emotional voice, while MGTs tend to maintain a neutral, impersonal tone, avoiding subjective stance or interpersonal language.

\paragraph{Structural or formal patterning.} This refers to the overall organization and formatting of a text, such as the use of lists, repeated sentence structures, or rigid paragraph symmetry. MGTs often rely on highly formulaic patterns and structural templates, whereas HGTs usually follow more flexible and varied writing conventions.

\paragraph{Content depth and specificity.} This refers to the degree of detail, conceptual clarity, and domain-specific insight within a text. HGTs tend to provide richer context and nuanced discussion, while MGTs often remain general, vague, or overly simplistic due to limited reasoning depth or training data generalization.

\paragraph{Personal or narrative elements.} This refers to the presence of storytelling, analogies, humor, or personal reflection. HGTs frequently include such elements to convey experience or emotion, whereas MGTs typically avoid subjective expression and may struggle to produce authentic narrative voice.

\paragraph{Bias.} This indicates the presence of prejudicial or favoring tendencies in a text. HGTs are more likely to reflect personal or societal biases, while MGTs generally show fewer biases, though they can still mirror biases present in their training data.

These categories helped our human annotators to be more certain on some HGTs and MGTs. However, texts lacking definitive features were categorized as ``undecided'' based on the absence of clear human or machine indicators.

\section{Further Discussions}

\subsection{Justification for Ternary Classification}

The introduction of the ``undecided'' category has sparked a considerable debate concerning its validity. For instance, a text in Table~\ref{tab:example} was categorized as ``undecided'' by our human annotators, whereas detection tools like GPTZero, Sapling, and Binoculars identified it as MGT -- a classification that is technically correct. However, according to our human annotators, these texts were aptly placed in the ``undecided'' category, arguing that there was no definitive reason to label them strictly as MGTs, suggesting instances where LLMs might merely be mimicking human-like output. More examples of this kind can be found in Appendix~\ref{appendix:binary_vs_ternary_more_examples}.

Upon reviewing the explanation notes provided by our human annotators, we observed that characteristics of MGTs and HGTs often overlap across several categories. This overlap creates a gray area in determining the origin of the text, as the boundaries between MGT and HGT are not always clear-cut. Moreover, since MGTs are trained on and derive from HGTs, they can produce texts that are indistinguishable from human writings.

Although it is apparent that human annotators struggled with accurately distinguishing the ``undecided'' category from the other two, this ambiguity also underscores the complexity of text generation origins. Despite these challenges, the ternary classification provides a framework that can guide further refinement in identifying and differentiating these text categories. Future efforts should focus on establishing more precise criteria to discern the unique characteristics and distinctions among these three labels.

\subsection{Explainability of Detectors}

In our recent experiments, human annotators categorized texts into three groups and provided explanation notes for their classifications. The types of explanation notes identified align with findings in past research, highlighting key factors like errors, perplexity, repetition, and readability as crucial in distinguishing between MGTs and HGTs. For instance, studies such as those by~\citet{mindner2023classification} and~\citet{munoz2023contrasting} have documented similar observations regarding language usage differences between MGTs and HGTs.

Human annotators' explanation notes are predominantly qualitative, yet quantitative measures can also be applied, particularly for aspects like spelling and grammatical errors, perplexity, and readability. For instance, tools such as Grammarly can assist in evaluating spelling and grammatical errors, while NLP tools can be used to calculate text perplexity. Readability can be assessed using existing formalas such as the Flesch Reading Ease~\cite{Flesch1948readability} and Flesch-Kincaid Grade Level~\cite{Kincaid1975readbility}. Our experiments demonstrate a gradual decline in readability and perplexity scores from texts in the ``human'' category to the ``undecided'' category, and finally to the ``machine'' category. More detailed experimental results can be found in Appendix~\ref{appendix:quantitative_human_explanations}.

Despite the robustness of human explanations, which are grounded in common sense and supported by the literature, discrepancies still exist. For example, \citet{Hans2024Binoculars} introduced the ``capybara problem'', where both prompts and responses with high perplexity can lead to misjudgments about text origin, both by humans and automated detectors, particularly when prompt details are unknown. Addressing the ``capybara problem'' involves creating prompts that encourage LLMs to produce features typical of HGTs, as detailed by our annotators. Effective strategies for this are outlined in Appendix~\ref{appendix:human_explanations_counterexamples}. 
% Moreover, advancements in LLMs like the reduction of unnecessary repetition from ChatGPT-3.5 to ChatGPT-4 demonstrate ongoing improvements, as discussed in Appendix~\ref{appendix:improvements_ChatGPT-3.5to4}.

Currently, detector explainability is very limited, and there are instances where provided explanations do not accurately reflect the underlying reasoning of decisions. Future research should aim to enhance the credibility and transparency of detectors by incorporating explainability modules or integrating explainable AI (XAI) components into existing and future MGT detectors.

Future studies should also focus on a better understanding of the nuances between HGTs and MGTs, possibly through user studies that assess perception and comprehension. Technologically, efforts could be directed towards improving the user interfaces of MGT detectors to provide more user-friendly explanations, potentially in an interactive, personalized and contextualized manner. For example, models could indicate whether sentences are derived from what training data or newly generated, potentially using a confidence scale to differentiate between entirely new creations and slight modifications of existing data. Such transparency could greatly enhance the explainability of AI-generated content.

\section{Conclusion}

This paper explores the effectiveness and challenges associated with current text detection systems. We initially set up a binary classification experiment to identify the top-performing detectors and LLMs that excel in resisting such top-performing detectors. The study was then extended to include a ternary classification framework involving datasets from ChatGPT-4, ChatGPT-3.5, and human sources, where human annotators assessed and explained their classification decisions. The results affirm the relevance of our ternary classification approach, particularly as LLMs continue to advance and produce increasingly human-like texts, making traditional binary classification approaches less meaningful. Our analysis indicates that while current detectors are lacking in explainability, the insights provided by human annotators are valuable for guiding future researcher on MGT detection. These outcomes lead us to recommend enhancements for future detection systems and their explanatory components.

\newpage

\section*{Limitations}

This study is subject to several limitations. While the human-produced explanations from our study contribute valuable perspectives, they predominantly serve as recommendations and pointers for further research on improving detection systems. Additionally, given the ongoing advancements in LLM technology, new research opportunities and directions are likely to emerge, necessitating continual updates and revisions to our approach.

\section*{Ethics Statements}

All experiments were conducted using publicly available LLMs and datasets. For the datasets we constructed as part of this work, no personal or private information was included. Human annotation was carried out by recruited contributors who are not co-authors of this paper. The annotation procedure was approved through an institutional research ethics review. More details about the annotation process can be found in Appendix~\ref{appendix:human_annotation}.

\bibliography{reference}

\begin{thebibliography}{46}
\expandafter\ifx\csname natexlab\endcsname\relax\def\natexlab#1{#1}\fi

\bibitem[{Abdalla et~al.(2023)Abdalla, Malberg, Dementieva, Mosca, and Groh}]{abdalla2023benchmark}
Mohamed Hesham~Ibrahim Abdalla, Simon Malberg, Daryna Dementieva, Edoardo Mosca, and Georg Groh. 2023.
\newblock \href {https://doi.org/10.3390/info14100522} {A benchmark dataset to distinguish human-written and machine-generated scientific papers}.
\newblock \emph{Information}, 14(10):522:1--522:33.

\bibitem[{Andr{\'e} et~al.(2023)Andr{\'e}, Eriksen, Jakobsen, Mingolla, and Thomsen}]{andre2023detecting}
Christopher M.~J. Andr{\'e}, Helene F.~L. Eriksen, Emil~J. Jakobsen, Luca C.~B. Mingolla, and Nicolai~B. Thomsen. 2023.
\newblock \href {https://ceur-ws.org/Vol-3551/paper3.pdf} {Detecting {AI} authorship: Analyzing descriptive features for {AI} detection}.
\newblock In \emph{Proceedings of the 7th Workshop on Natural Language for Artificial Intelligence (NL4AI 2023)}.

\bibitem[{Antoun et~al.(2023)Antoun, Sagot, and Seddah}]{Antoun2024text2source}
Wissam Antoun, Beno{\^i}t Sagot, and Djam{\'e} Seddah. 2023.
\newblock \href {https://aclanthology.org/2024.lrec-main.665} {From text to source: Results in detecting large language model-generated content}.
\newblock In \emph{Proceedings of the 2024 Joint International Conference on Computational Linguistics, Language Resources and Evaluation (LREC-COLING 2024)}, pages 7531--7543.

\bibitem[{Bao et~al.(2024)Bao, Zhao, Teng, Yang, and Zhang}]{Bao2024Fast-DetectGPT}
Guangsheng Bao, Yanbin Zhao, Zhiyang Teng, Linyi Yang, and Yue Zhang. 2024.
\newblock \href {https://openreview.net/forum?id=Bpcgcr8E8Z} {{Fast-DetectGPT}: Efficient zero-shot detection of machine-generated text via conditional probability curvature}.
\newblock In \emph{Proceedings of the 12th International Conference on Learning Representations}.

\bibitem[{Bhattacharjee et~al.(2023)Bhattacharjee, Kumarage, Moraffah, and Liu}]{Bhattacharjee2023ConDA}
Amrita Bhattacharjee, Tharindu Kumarage, Raha Moraffah, and Huan Liu. 2023.
\newblock \href {https://doi.org/10.18653/v1/2023.ijcnlp-main.40} {{ConDA}: Contrastive domain adaptation for {AI}-generated text detection}.
\newblock In \emph{Proceedings of the 13th International Joint Conference on Natural Language Processing and the 3rd Conference of the Asia-Pacific Chapter of the Association for Computational Linguistics (Volume 1: Long Papers)}, pages 598--610.

\bibitem[{Breiman et~al.(1984)Breiman, Friedman, Olshen, and Stone}]{Breiman1984}
Leo Breiman, Jerome Friedman, Richard Olshen, and Charles Stone. 1984.
\newblock \href {https://doi.org/10.1201/9781315139470} {\emph{Classification and Regression Trees}}.
\newblock Chapman and Hall/CRC.

\bibitem[{{Copyleaks Technologies Ltd.}(2023)}]{copyleaks_website}
{Copyleaks Technologies Ltd.} 2023.
\newblock \href {https://copyleaks.com/ai-content-detector} {{AI} detector | {ChatGPT} detector | {AI} checker - {CopyLeaks}}.
\newblock Online.

\bibitem[{Cortes and Vapnik(1995)}]{Cortes1995SVN}
Corinna Cortes and Vladimir Vapnik. 1995.
\newblock \href {https://doi.org/10.1007/BF00994018} {Support-vector networks}.
\newblock \emph{Machine Learning}, 20(3):273--297.

\bibitem[{Cox(1958)}]{Cox1958}
D.~R. Cox. 1958.
\newblock \href {https://doi.org/10.1111/j.2517-6161.1958.tb00292.x} {The regression analysis of binary sequences}.
\newblock \emph{Journal of the Royal Statistical Society. Series B (Methodological)}, 20(2):215--242.

\bibitem[{Dubey et~al.(2024)Dubey, Jauhri, Pandey, Kadian, Al-Dahle, Letman, Mathur, Schelten, Yang, Fan et~al.}]{dubey2024llama}
Abhimanyu Dubey, Abhinav Jauhri, Abhinav Pandey, Abhishek Kadian, Ahmad Al-Dahle, Aiesha Letman, Akhil Mathur, Alan Schelten, Amy Yang, Angela Fan, et~al. 2024.
\newblock The llama 3 herd of models.
\newblock \emph{arXiv preprint arXiv:2407.21783}.

\bibitem[{Fan et~al.(2019)Fan, Jernite, Perez, Grangier, Weston, and Auli}]{Fan2019ELI5}
Angela Fan, Yacine Jernite, Ethan Perez, David Grangier, Jason Weston, and Michael Auli. 2019.
\newblock \href {https://doi.org/10.48550/arXiv.1907.09190} {{ELI5}: Long form question answering}.
\newblock arXiv:1907.09190 [cs.CL].

\bibitem[{Fleiss(1971)}]{Fleiss1971kappa}
Joseph~L. Fleiss. 1971.
\newblock \href {https://doi.org/10.1037/h0031619} {Measuring nominal scale agreement among many raters}.
\newblock \emph{Psychological Bulletin}, 76(5):378--382.

\bibitem[{Flesch(1948)}]{Flesch1948readability}
Rudolf Flesch. 1948.
\newblock \href {https://doi.org/10.1037/h0057532} {A new readability yardstick}.
\newblock \emph{Journal of Applied Psychology}, 32(3):221--233.

\bibitem[{Gehrmann et~al.(2019)Gehrmann, Strobelt, and Rush}]{gehrmann2019gltr}
Sebastian Gehrmann, Hendrik Strobelt, and Alexander~M. Rush. 2019.
\newblock \href {https://doi.org/10.48550/arXiv.1906.04043} {{GLTR}: Statistical detection and visualization of generated text}.
\newblock arXiv:1906.04043 [cs.CL].

\bibitem[{Guo et~al.(2023)Guo, Zhang, Wang, Jiang, Nie, Ding, Yue, and Wu}]{Guo2023ChatGPT_vs_humans}
Biyang Guo, Xin Zhang, Ziyuan Wang, Minqi Jiang, Jinran Nie, Yuxuan Ding, Jianwei Yue, and Yupeng Wu. 2023.
\newblock \href {https://doi.org/10.48550/arXiv.2301.07597} {How close is {ChatGPT} to human experts? comparison corpus, evaluation, and detection}.
\newblock arXiv:2301.07597 [cs.CL].

\bibitem[{Hans et~al.(2024)Hans, Schwarzschild, Cherepanova, Kazemi, Saha, Goldblum, Geiping, and Goldstein}]{Hans2024Binoculars}
Abhimanyu Hans, Avi Schwarzschild, Valeriia Cherepanova, Hamid Kazemi, Aniruddha Saha, Micah Goldblum, Jonas Geiping, and Tom Goldstein. 2024.
\newblock \href {https://doi.org/10.48550/arXiv.2401.12070} {Spotting {LLMs} with {Binoculars}: Zero-shot detection of machine-generated text}.
\newblock arXiv:2401.12070 [cs.CL].

\bibitem[{Hassabis and the Gemini~Team(2023)}]{gemini2023}
Demis Hassabis and the Gemini~Team. 2023.
\newblock \href {https://blog.google/technology/ai/introducing-gemini/} {Introducing {Gemini}: {Google}'s most capable {AI} model yet}.
\newblock Online.

\bibitem[{He et~al.(2023)He, Shen, Chen, Backes, and Zhang}]{He2023MGTBench}
Xinlei He, Xinyue Shen, Zeyuan Chen, Michael Backes, and Yang Zhang. 2023.
\newblock \href {https://doi.org/10.48550/arXiv.2303.14822} {{MGTBench}: Benchmarking machine-generated text detection}.
\newblock arXiv:2303.14822 [cs.CR].

\bibitem[{{Inspera}(2023)}]{crossplag_website}
{Inspera}. 2023.
\newblock \href {https://crossplag.com/ai-content-detector/} {{AI} content detector}.
\newblock Online.

\bibitem[{Joshi et~al.(2024)Joshi, Pocker, Dandekar, Dandekar, and Panat}]{Joshi2024HULLMI}
Prathamesh~Dinesh Joshi, Sahil Pocker, Raj~Abhijit Dandekar, Rajat Dandekar, and Sreedath Panat. 2024.
\newblock \href {https://arxiv.org/abs/2409.04808} {Hullmi: Human vs llm identification with explainability}.
\newblock \emph{arXiv preprint arXiv:2409.04808}.

\bibitem[{Kincaid et~al.(1975)Kincaid, Fishburne, Rogers, and Chissom}]{Kincaid1975readbility}
J.~Peter Kincaid, Robert P.~Jr Fishburne, Richard~L. Rogers, and Brad~S. Chissom. 1975.
\newblock \href {https://stars.library.ucf.edu/istlibrary/56/} {Derivation of new readability formulas ({Automated Readability Index}, {Fog Count} and {Flesch Reading Ease Formula}) for navy enlisted personnel}.
\newblock Technical Report Research Branch Report 8-75, Naval Technical Training Command Millington TN Research Branch.

\bibitem[{Kumari et~al.(2024)Kumari, Pegoraro, Fereidooni, and Sadeghi}]{Kumari2024DEMASQ}
Kavita Kumari, Alessandro Pegoraro, Hossein Fereidooni, and Ahmad-Reza Sadeghi. 2024.
\newblock \href {https://www.ndss-symposium.org/ndss-paper/demasq-unmasking-the-chatgpt-wordsmith/} {{DEMASQ}: Unmasking the {ChatGPT} wordsmith}.

\bibitem[{Lundberg and Lee(2017)}]{lundberg2017unified}
Scott~M. Lundberg and Su-In Lee. 2017.
\newblock \href {https://proceedings.neurips.cc/paper_files/paper/2017/file/8a20a8621978632d76c43dfd28b67767-Paper.pdf} {A unified approach to interpreting model predictions}.
\newblock \emph{Advances in Neural Information Processing Systems}, 30:4768--4777.

\bibitem[{Ma and Wang(2024)}]{Ma2024TOCSIN}
Shixuan Ma and Quan Wang. 2024.
\newblock \href {https://aclanthology.org/2024.emnlp-main.971/} {Zero-shot detection of llm-generated text using token cohesiveness}.
\newblock \emph{Proceedings of the 2024 Conference on Empirical Methods in Natural Language Processing (EMNLP)}.

\bibitem[{Mindner et~al.(2023)Mindner, Schlippe, and Schaaff}]{mindner2023classification}
Lorenz Mindner, Tim Schlippe, and Kristina Schaaff. 2023.
\newblock \href {https://doi.org/10.1007/978-981-99-7947-9_12} {Classification of human-and {AI}-generated texts: Investigating features for {ChatGPT}}.
\newblock In \emph{Artificial Intelligence in Education Technologies: New Development and Innovative Practices -- Proceedings of 2023 4th International Conference on Artificial Intelligence in Education Technology}, pages 152--170. Springer.

\bibitem[{Mitchell et~al.(2023)Mitchell, Lee, Khazatsky, Manning, and Finn}]{Mitchell2023DetectGPT}
Eric Mitchell, Yoonho Lee, Alexander Khazatsky, Christopher~D. Manning, and Chelsea Finn. 2023.
\newblock \href {https://proceedings.mlr.press/v202/mitchell23a.html} {{DetectGPT}: Zero-shot machine-generated text detection using probability curvature}.
\newblock In \emph{Proceedings of the 40th International Conference on Machine Learning}, pages 24950--24962.

\bibitem[{Mitrovi{\'c} et~al.(2023)Mitrovi{\'c}, Andreoletti, and Ayoub}]{mitrovic2023chatgpt}
Sandra Mitrovi{\'c}, Davide Andreoletti, and Omran Ayoub. 2023.
\newblock \href {https://doi.org/10.48550/arXiv.2301.13852} {{ChatGPT} or human? detect and explain. explaining decisions of machine learning model for detecting short {ChatGPT}-generated text}.
\newblock arXiv:2301.13852 [cs.CL].

\bibitem[{Mu{\~n}oz-Ortiz et~al.(2023)Mu{\~n}oz-Ortiz, G{\'o}mez-Rodr{\'\i}guez, and Vilares}]{munoz2023contrasting}
Alberto Mu{\~n}oz-Ortiz, Carlos G{\'o}mez-Rodr{\'\i}guez, and David Vilares. 2023.
\newblock \href {https://doi.org/10.48550/arXiv.2308.09067} {Contrasting linguistic patterns in human and {LLM}-generated text}.
\newblock arXiv:2308.09067 [cs.CL].

\bibitem[{{OpenAI}(2021)}]{openai_text_classifier}
{OpenAI}. 2021.
\newblock \href {https://github.com/openai/gpt-2-output-dataset/} {{GPT}-2 output detector demo}.
\newblock Online.

\bibitem[{{OpenAI}(2023)}]{OpenAI2023GTP4}
{OpenAI}. 2023.
\newblock \href {https://www.openai.com/research/gpt-4} {{GPT-4}}.
\newblock Online.

\bibitem[{{OpenAI}(2024)}]{OpenAI2024GTP4O}
{OpenAI}. 2024.
\newblock \href {https://openai.com/index/gpt-4o-system-card/} {{GPT-4O}}.
\newblock Online.

\bibitem[{{Originality.AI}(2024)}]{originality_ai_website}
{Originality.AI}. 2024.
\newblock \href {https://app.originality.ai/api-access} {{Originality.AI}}.
\newblock Online.

\bibitem[{Rezaei et~al.(2024)Rezaei, Kwon, Sanayei, Singh, and Bethard}]{Rezaei2024CLULab}
Mohammadhossein Rezaei, Yeaeun Kwon, Reza Sanayei, Abhyuday Singh, and Steven Bethard. 2024.
\newblock \href {https://aclanthology.org/2024.semeval-1.215/} {Clulab-uofa at semeval-2024 task 8: Detecting machine-generated text using triplet-loss-trained text similarity and text classification}.
\newblock \emph{Proceedings of the 18th International Workshop on Semantic Evaluation (SemEval-2024)}.

\bibitem[{Rosenblatt(1958)}]{Rosenblatt1958perceptron}
Frank Rosenblatt. 1958.
\newblock \href {https://doi.org/10.1037/h0042519} {The perceptron: A probabilistic model for information storage and organization in the brain}.
\newblock \emph{Psychological Review}, 65(6):386--408.

\bibitem[{{Sapling AI Team}(2023)}]{sapling_website}
{Sapling AI Team}. 2023.
\newblock \href {https://sapling.ai/ai-content-detector} {{Sapling}}.
\newblock Online.

\bibitem[{Solaiman et~al.(2019)Solaiman, Brundage, Clark, Askell, Herbert-Voss, Wu, Radford, Krueger, Kim, Kreps, McCain, Newhouse, Blazakis, McGuffie, and Wang}]{solaiman2019release}
Irene Solaiman, Miles Brundage, Jack Clark, Amanda Askell, Ariel Herbert-Voss, Jeff Wu, Alec Radford, Gretchen Krueger, Jong~Wook Kim, Sarah Kreps, Miles McCain, Alex Newhouse, Jason Blazakis, Kris McGuffie, and Jasmine Wang. 2019.
\newblock \href {https://doi.org/10.48550/arXiv.1908.09203} {Release strategies and the social impacts of language models}.
\newblock arXiv:1908.09203 [cs.CL].

\bibitem[{Su et~al.(2023)Su, Zhuo, Wang, and Nakov}]{su2023detectllm}
Jinyan Su, Terry~Yue Zhuo, Di~Wang, and Preslav Nakov. 2023.
\newblock \href {https://doi.org/10.48550/arXiv.2306.05540} {{DetectLLM}: Leveraging log rank information for zero-shot detection of machine-generated text}.
\newblock arXiv:2306.05540 [cs.CL].

\bibitem[{Tian et~al.(2023)Tian, Cui, Kusio et~al.}]{gptzero_website}
Edward Tian, Alex Cui, Olivia Kusio, et~al. 2023.
\newblock \href {https://www.gptzero.me/} {{GPTZero}}.
\newblock Online.

\bibitem[{Wang et~al.(2023)Wang, Li, Ren, Jiang, Zhang, and Qiu}]{Wang2023SeqXGPT}
Pengyu Wang, Linyang Li, Ke~Ren, Botian Jiang, Dong Zhang, and Xipeng Qiu. 2023.
\newblock \href {https://aclanthology.org/2023.emnlp-main.73} {{SeqXGPT}: Sentence-level {AI}-generated text detection}.
\newblock In \emph{Proceedings of the 2023 Conference on Empirical Methods in Natural Language Processing}, pages 1144--1156.

\bibitem[{Wu et~al.(2024)Wu, Yang, Zhan, Yuan, Wong, and Chao}]{Wu2024survey}
Junchao Wu, Shu Yang, Runzhe Zhan, Yulin Yuan, Derek~F. Wong, and Lidia~S. Chao. 2024.
\newblock \href {https://doi.org/10.48550/arXiv.2310.14724} {A survey on {LLM}-generated text detection: Necessity, methods, and future directions}.
\newblock arXiv:2310.14724v3 [cs.CL].

\bibitem[{Wu et~al.(2023)Wu, Pang, Shen, Cheng, and Chua}]{Wu2023LLMDet}
Kangxi Wu, Liang Pang, Huawei Shen, Xueqi Cheng, and Tat-Seng Chua. 2023.
\newblock \href {https://doi.org/10.18653/v1/2023.findings-emnlp.139} {{LLMDet}: A third party large language models generated text detection tool}.
\newblock In \emph{Findings of the Association for Computational Linguistics: EMNLP 2023}, pages 2113--2133.

\bibitem[{Wu and Xiang(2023)}]{wu2023mfd}
Zhendong Wu and Hui Xiang. 2023.
\newblock \href {https://doi.org/10.21203/rs.3.rs-3226684/v1} {{MFD}: Multi-feature detection of {LLM}-generated text}.

\bibitem[{Yang et~al.(2024)Yang, Yang, Zhang, Hui, Zheng, Yu, Li, Liu, Huang, Wei et~al.}]{yang2024qwen2}
An~Yang, Baosong Yang, Beichen Zhang, Binyuan Hui, Bo~Zheng, Bowen Yu, Chengyuan Li, Dayiheng Liu, Fei Huang, Haoran Wei, et~al. 2024.
\newblock Qwen2. 5 technical report.
\newblock \emph{arXiv preprint arXiv:2412.15115}.

\bibitem[{Yang et~al.(2023)Yang, Pan, Zhao, Chen, Petzold, Wang, and Cheng}]{Yang2023a}
Xianjun Yang, Liangming Pan, Xuandong Zhao, Haifeng Chen, Linda Petzold, William~Yang Wang, and Wei Cheng. 2023.
\newblock \href {https://doi.org/10.48550/arXiv.2310.15654} {A survey on detection of {LLMs}-generated content}.
\newblock arXiv:2310.15654 [cs.CL].

\bibitem[{{ZeroGPT.com}(2023)}]{zerogpt_website}
{ZeroGPT.com}. 2023.
\newblock \href {https://www.zerogpt.com/} {{ZeroGPT}}.
\newblock Online.

\bibitem[{Zhang et~al.(2024)Zhang, Song, Yang, Li, Han, and Tan}]{zhang2024detecting}
Shuhai Zhang, Yiliao Song, Jiahao Yang, Yuanqing Li, Bo~Han, and Mingkui Tan. 2024.
\newblock \href {https://openreview.net/forum?id=3fEKavFsnv} {Detecting machine-generated texts by multi-population aware optimization for maximum mean discrepancy}.
\newblock In \emph{Proceedings of The 12th International Conference on Learning Representations}.

\end{thebibliography}

\appendix

\section{Detailed Results of Binary Classification Experiments}
\label{appendix:binary_performance_results}

%Table~\ref{tab:table3}, Table~\ref{tab:table4} and Table~\ref{tab:table5} presents detailed results data for various detectors in binary classification tests across different datasets.

In the binary classification experiments, the performance of various detectors on datasets consisting of texts generated by GPT-4o and humans is presented in Table~\ref{tab:binary_performance_GPT4o}. Tables~\ref{tab:binary_performance_Gemini_Pro}, \ref{tab:binary_performance_LLaMA3.3_70B} and \ref{tab:binary_performance_Qwen2_72B} show the specific performance of different detectors on texts generated by ChatGPT-3.5, LLaMA-13B, and Gemini Pro, respectively.

\begin{table*}[h]
\centering
\small
\begin{tabularx}{\textwidth}{@{}l *{8}{>{\centering\arraybackslash}X} @{}}
\toprule
\multirow{2}{*}{Models} & \multirow{2}{*}{Accuracy} & \multicolumn{3}{c}{Machine as Positive} & \multicolumn{3}{c}{Human as Positive} & \multirow{2}{*}{Macro F1} \\
\cmidrule(lr{0.5em}){3-5} \cmidrule(lr{0.5em}){6-8}
& & Precision & Recall & F1 & Precision & Recall & F1 & \\
\midrule
GPTZero & 94.95\% & 93.01\% & 97.20\% & 95.06\% & 97.07\% & 92.70\% & 94.83\% & 94.95\% \\
Sapling & 93.20\% & 89.93\% & 97.30\% & 93.47\% & 97.06\% & 89.10\% & 92.91\% & 93.19\% \\
Binoculars & 86.80\% & 80.36\% & 97.40\% & 88.07\% & 96.70\% & 76.20\% & 85.23\% & 86.65\% \\
Fast-DetectGPT & 85.20\% & 89.02\% & 80.30\% & 84.44\% & 82.06\% & 90.10\% & 85.89\% & 85.16\% \\
MMD-MP & 81.25\% & 96.71\% & 64.70\% & 77.53\% & 73.48\% & 97.80\% & 83.91\% & 80.72\% \\
DEMASQ & 60.80\% & 57.07\% & 87.20\% & 68.99\% & 72.88\% & 34.40\% & 46.74\% & 57.86\% \\
DetectGPT & 69.05\% & 93.59\% & 40.90\% & 56.92\% & 62.19\% & 97.20\% & 75.85\% & 66.39\% \\
\bottomrule
\end{tabularx}
\caption{Binary classification performance of different detectors on the dataset of Gemini Pro}
\label{tab:binary_performance_Gemini_Pro}
\end{table*}

\begin{table*}[h]
\centering
\small
\begin{tabularx}{\textwidth}{@{}l *{8}{>{\centering\arraybackslash}X} @{}}
\toprule
\multirow{2}{*}{Models} & \multirow{2}{*}{Accuracy} & \multicolumn{3}{c}{Machine as Positive} & \multicolumn{3}{c}{Human as Positive} & \multirow{2}{*}{Macro F1} \\
\cmidrule(lr{0.5em}){3-5} \cmidrule(lr{0.5em}){6-8}
& & Precision & Recall & F1 & Precision & Recall & F1 & \\
\midrule
GPTZero & 93.15\% & 92.77\% & 93.60\% & 93.18\% & 93.54\% & 92.70\% & 93.12\% & 93.15\% \\
Sapling &93.60\% & 90.00\% & 98.10\% & 93.88\% & 97.91\% & 89.10\% & 93.30\% & 93.59\% \\
Binoculars & 87.55\% & 80.60\% & 98.90\% & 88.82\% & 98.58\% & 76.20\% & 85.96\% & 87.39\% \\
Fast-DetectGPT & 83.25\% & 88.53\% & 76.40\% & 82.02\% & 79.24\% & 90.10\% & 84.32\% & 83.17\% \\
MMD-MP & 83.55\% & 96.92\% & 69.30\% & 80.82\% & 76.11\% & 97.80\% & 85.60\% & 83.21\% \\
DEMASQ & 62.70\% & 58.11\% & 91.00\% & 70.93\% & 79.26\% & 34.40\% & 47.98\% & 59.45\% \\
DetectGPT & 65.55\% & 92.37\% & 33.90\% & 49.60\% & 59.52\% & 97.20\% & 73.83\% & 61.71\% \\
\bottomrule
\end{tabularx}
\caption{Binary classification performance of different detectors on the dataset of LLaMA3.3-70B}
\label{tab:binary_performance_LLaMA3.3_70B}
\end{table*}

\begin{table*}[h]
\centering
\small
\begin{tabularx}{\textwidth}{@{}l *{8}{>{\centering\arraybackslash}X} @{}}
\toprule
\multirow{2}{*}{Models} & \multirow{2}{*}{Accuracy} & \multicolumn{3}{c}{Machine as Positive} & \multicolumn{3}{c}{Human as Positive} & \multirow{2}{*}{Macro F1} \\
\cmidrule(lr{0.5em}){3-5} \cmidrule(lr{0.5em}){6-8}
& & Precision & Recall & F1 & Precision & Recall & F1 & \\
\midrule
GPTZero & 92.05\% & 92.60\% & 91.40\% & 92.00\% & 91.51\% & 92.70\% & 92.10\% & 92.05\% \\
Sapling & 92.90\% & 89.87\% & 96.70\% & 93.16\% & 96.43\% & 89.10\% & 92.62\% & 92.89\% \\
Binoculars & 85.70\% & 80.00\% & 95.20\% & 86.94\% & 94.07\% & 76.20\% & 84.20\% & 85.57\% \\
Fast-DetectGPT & 78.00\% & 86.94\% & 65.90\% & 74.97\% & 72.54\% & 90.10\% & 80.37\% & 77.67\% \\
MMD-MP & 76.05\% & 96.11\% & 54.30\% & 69.39\% & 68.15\% & 97.80\% & 80.33\% & 74.86\% \\
DEMASQ & 62.50\% & 58.00\% & 90.60\% & 70.73\% & 78.54\% & 34.40\% & 47.84\% & 59.29\% \\
DetectGPT & 55.35\% & 82.82\% & 13.50\% & 23.22\% & 52.91\% & 97.20\% & 68.52\% & 45.87\% \\
\bottomrule
\end{tabularx}
\caption{Binary classification performance of different detectors on the dataset of Qwen2-72B}
\label{tab:binary_performance_Qwen2_72B}
\end{table*}

\section{More about Explanatory Power of the Six Metrics of GPTZero}
\label{appendix:GPTZero_explainability}

In the explanations provided by GPTZero, six explainability metrics are identified: Readability, Percent SAT, Simplicity, Perplexity, Burstiness, and Average Sentence Length. For all texts evaluated by GPTZero and their corresponding six feature values, we created a new dataset to analyze the explainability provided by GPTZero. The ground truth is based on GPTZero's evaluation results. We partitioned the dataset into training and test sets with an 8:2 ratio. We trained four classifiers: Logistic Regression, SVC, Perceptron, and Decision Tree. The weights and accuracy of the different features obtained from these classifiers are presented in Table~\ref{tab:tableb1}. From the weights, it is evident that the two most effective metrics in GPTZero's explainability are perplexity and readability scores. The remaining metrics contribute minimally to the final results. Additionally, the trained classifier exhibits relatively low accuracy, suggesting that GPTZero employs more complex calculations or utilizes additional sophisticated features that are not disclosed.

\begin{table*}[h!]
\centering
\small
\begin{tabular}{@{}c*{6}{>{\centering\arraybackslash}p{1.2cm}}c@{}}
\toprule
\multirow{2}{*}{\textbf{Classifier}} & \multicolumn{6}{c}{\textbf{Feature Importances}} & \textbf{Accuracy (\%)} \\
\cmidrule{2-8}
 & \textbf{Readability} & \textbf{PSAT} & \textbf{Simplicity} & \textbf{Perplexity} & \textbf{Burstiness} & \textbf{ASL} & \\
\midrule
LR & 3.094 & -0.857 & 1.821 & -2.517 & 0.036 & 0.713 & 75.76 \\
SVC & 2.637 & -0.671 & 2.677 & -2.189 & 0.051 & 0.654 & 77.27 \\
Perceptron & 4.109 & -0.991 & 8.148 & -4.437 & 0.417 & 1.039 & 78.79 \\
Decision Tree & 0.289 & 0.016 & 0.199 & 0.205 & 0.183 & 0.109 & 75.76 \\
\bottomrule
\end{tabular}
\caption{Weights and accuracy of different classifiers using GPTZero's six explainability metrics as features. LR stands for Logistic Regression. PSAT stands for Percent SAT. ASL stands for Average Sentence Length.}
\label{tab:tableb1}
\end{table*}

\section{Examples of Different Types of Explanations Given by Human Annotators}
\label{appendix:human_explanation_categories}

Our revised analysis categorized human annotators’ explanations into six main types. Below we present representative examples and analyses for each category to illustrate how annotators use specific textual features to inform their classification judgments.

\paragraph{Linguistic fluency and coherence.} In the text in Table~\ref{tab:tablec1}, grammatical inconsistencies, awkward syntax, and unusual punctuation patterns led annotators to classify the text as human-written. These traits suggest natural writing imperfection and reduced language monitoring, common in informal human communication.

\begin{table*}[h]
\centering
\small
\begin{tabular}{@{}p{\linewidth}@{}}
\toprule
\textbf{Text}: Quite difficult to follow. A single long paragraph without much sense of structure. Odd layout and some errors.\\
\bottomrule
\end{tabular}
\caption{An example of human annotators using linguistic fluency and coherence to judge authorship. This text is human-written, and the annotators labeled it ``human''.}
\label{tab:tablec1}
\end{table*}

\paragraph{Stylistic register and tone.} The text in Table~\ref{tab:tablec2} uses informal and emotionally expressive language such as “pretend you’re sculpting clay” and direct address (“you”), which suggests a conversational and subjective tone typical of human writing.

\begin{table*}[h]
\centering
\small
\begin{tabular}{@{}p{\linewidth}@{}}
\toprule
\textbf{Text}: Errors, colloquialisms, use of “you”. “Pretend you’re sculpting clay” sounds human.\\
\bottomrule
\end{tabular}
\caption{An example of stylistic register and tone guiding human judgment. The text is human-written and was labeled ``human''.}
\label{tab:tablec2}
\end{table*}

\paragraph{Structural or formal patterning.} The response in Table~\ref{tab:tablec3} is structurally rigid and formulaic, with repeated sentence patterns and list-based phrasing. These traits led annotators to label it as machine-generated.

\begin{table*}[h]
\centering
\small
\begin{tabular}{@{}p{\linewidth}@{}}
\toprule
\textbf{Text}: Too formulaically structured to be human. The explanation is basically a list of facts with headings.\\
\bottomrule
\end{tabular}
\caption{An example of structural/formal cues leading to a machine classification. This text is machine-generated and was labeled ``machine''.}
\label{tab:tablec3}
\end{table*}

\paragraph{Content depth and specificity.} The text in Table~\ref{tab:tablec4} was flagged as vague and overly general, lacking topic-specific insight. Annotators viewed this lack of depth as indicative of AI generation.

\begin{table*}[h]
\centering
\small
\begin{tabular}{@{}p{\linewidth}@{}}
\toprule
\textbf{Text}: It’s the list of themes. It feels mechanical and overly systematic. And it's very vague and general.\\
\bottomrule
\end{tabular}
\caption{An example of annotators using content specificity as an indicator. This text is machine-generated and labeled ``machine''.}
\label{tab:tablec4}
\end{table*}

\paragraph{Personal or narrative elements.} The text in Table~\ref{tab:tablec5} includes metaphorical language and subjective reflection. Annotators noted the analogy (“like sculpting clay”) as a human-like creative touch, contributing to their “human” label.

\begin{table*}[h]
\centering
\small
\begin{tabular}{@{}p{\linewidth}@{}}
\toprule
\textbf{Text}: The comparison with sculpting clay feels like the sort of comparison that would be made by a person who understands the tactile similarities.\\
\bottomrule
\end{tabular}
\caption{An example of personal or narrative elements used in human-authored text. Annotators labeled this as ``human''.}
\label{tab:tablec5}
\end{table*}

\paragraph{Bias.} The statement in Table~\ref{tab:tablec6} casually speculates on romantic interest in an offhand, possibly inappropriate manner. This reflects personal bias or humor, leading annotators to see it as human-written due to its subjectivity.

\begin{table*}[h]
\centering
\small
\begin{tabular}{@{}p{\linewidth}@{}}
\toprule
\textbf{Text}: That, or they just want to bone you.\\
\bottomrule
\end{tabular}
\caption{An example where personal/social bias is interpreted as evidence of human authorship. The text was written by a human and labeled ``human''.}
\label{tab:tablec6}
\end{table*}

\section{Counterexamples to Explanations Provided by Human Annotators}
\label{appendix:human_explanations_counterexamples}

Although human annotators often rely on specific cues, these cues are not always reliable. Here we present counterexamples for each explanation category, illustrating cases where the expected correlation between feature and authorship fails.

\paragraph{Linguistic fluency and coherence.} In Table~\ref{tab:tabled1}, the machine-generated text intentionally includes grammar and syntax errors (e.g., “I goes to school”) to simulate human imperfection, misleading annotators into labeling it as “human”.

\begin{table*}[h]
\centering
\small
\begin{tabular}{@{}p{\linewidth}@{}}
\toprule
\textbf{AI-generated text (with prompt to include grammar errors)}: Hello! My name is Li Wei. I lives in a small family. My father he is a teacher and my mother works in a hospital. I goes to school every day. I liking to read books and playing games after school...\\
\bottomrule
\end{tabular}
\caption{An AI-generated text with grammatical errors designed to mimic human mistakes.}
\label{tab:tabled1}
\end{table*}

\paragraph{Stylistic register and tone.} The AI-generated text in Table~\ref{tab:tabled2} uses casual tone and first-person voice (“I love my family”), which might be interpreted as human traits. However, it was generated by ChatGPT with stylistic prompts.

\begin{table*}[h]
\centering
\small
\begin{tabular}{@{}p{\linewidth}@{}}
\toprule
\textbf{AI-generated text (prompted to sound informal)}: I very love my family. We likes to go to the park. I happy to share about them.\\
\bottomrule
\end{tabular}
\caption{An example where AI mimics informal and personal tone. Annotators might misclassify this as human.}
\label{tab:tabled2}
\end{table*}

\paragraph{Structural or formal patterning.} The text in Table~\ref{tab:tabled3} follows a fragmented and inconsistent structure with weak transitions, yet it was generated by an AI system using low-cohesion prompts.

\begin{table*}[h]
\centering
\small
\begin{tabular}{@{}p{\linewidth}@{}}
\toprule
\textbf{AI-generated text (with low structure prompt)}: Machine learning is all about teaching computers to learn from data. Imagine a music app... Self-driving cars... Bias can be a problem...\\
\bottomrule
\end{tabular}
\caption{An AI-generated text with disrupted structure, simulating informal human writing.}
\label{tab:tabled3}
\end{table*}

\paragraph{Content depth and specificity.} Despite its rich vocabulary and abstract ideas, the text in Table~\ref{tab:tabled4} was generated by a language model. Annotators may misattribute such conceptual depth to human authorship.

\begin{table*}[h]
\centering
\small
\begin{tabular}{@{}p{\linewidth}@{}}
\toprule
\textbf{AI-generated text (high-perplexity prompt)}: In the penumbra of the quantum foam, time and space convolute into a symphony of probabilities...\\
\bottomrule
\end{tabular}
\caption{An AI-generated text with high conceptual density, potentially misclassified as human.}
\label{tab:tabled4}
\end{table*}

\paragraph{Personal or narrative elements.} The AI-written paragraph in Table~\ref{tab:tabled5} includes personal anecdotes and humor, but was generated by prompt engineering. Such surface-level narrative signals can mislead annotators.

\begin{table*}[h]
\centering
\small
\begin{tabular}{@{}p{\linewidth}@{}}
\toprule
\textbf{AI-generated text (prompted for personal tone)}: I watched it because I love Keanu Reeves. I should get the points for being the only guy with the balls to say I like this movie.\\
\bottomrule
\end{tabular}
\caption{An example of AI mimicking subjective narrative expression. Annotators may mislabel it as human.}
\label{tab:tabled5}
\end{table*}

\paragraph{Bias.} In Table~\ref{tab:tabled6}, the AI-generated text contains exaggerated and biased descriptions of environmental conditions. Despite being generated, its opinionated stance might be mistaken for human-authored satire or emotional expression.

\begin{table*}[h]
\centering
\small
\begin{tabular}{@{}p{\linewidth}@{}}
\toprule
\textbf{AI-generated text (bias prompt)}: The sanitary conditions are so deplorable that even wildlife struggles to survive...\\
\bottomrule
\end{tabular}
\caption{An AI-generated text that expresses subjective bias in a way similar to human satire.}
\label{tab:tabled6}
\end{table*}

\section{Examples of Changed Judgments in GPTZero Evaluations}
\label{sec:appendixf}

Regarding the test results of GPTZero versions from December 1, 2023, and May 1, 2024, the judgment outcomes for the two texts have changed. Both texts were machine-generated but were labeled as ``undecided'' by our human coders. Initially, GPTZero classified these texts as ``human'', but in the updated version, the classification has changed to ``AI''.

The feature values of the two texts in Tables~\ref{tab:tablef1} and \ref{tab:tablef2} remained completely consistent across both tests. However, the evaluation results were entirely opposite. This indicates that GPTZero operates with a more complex mechanism, and the explanations provided may not be highly interpretable.

\begin{table*}[h]
\centering
\small
\begin{tabular}{@{}p{\linewidth}@{}} % 使用p{width}指定每列宽度
\toprule
\textbf{Source}: ChatGPT-4

\textbf{Text}: Sweating itself does not directly cause colds. Colds are caused by viruses, not by being cold or sweating. However, if you sweat and then get chilled, this might weaken your immune system temporarily, making you more susceptible to catching a cold virus. Additionally, the belief that sweating leads to colds might stem from confusing the symptoms of a cold, which can include sweating, with the cause of the cold.\\
\midrule
\textbf{GPTZero result}: AI

\textbf{GPTZero explanations}: Readability: 61.8(Medium) Percent SAT: 2.5(Medium) Simplicity: 40.0(Medium) Perplexity: 34.4(Medium) Burstiness: 36.0(Medium) Average sentence length: 17.3(Medium)\\
\bottomrule
\end{tabular}
\caption{Comparison between abstract scores from GPTZero and human-readable explanations}
\label{tab:tablef1}
\end{table*}

\begin{table*}[h]
\centering
\small
\begin{tabular}{@{}p{\linewidth}@{}} % 使用p{width}指定每列宽度
\toprule
\textbf{Source}: ChatGPT-4

\textbf{Text}: "Imagine sending a toy camera tied to a super long string into a deep, dark well (like a black hole). If you try to pull it back, the string would probably break because the well is so strong it can even pull light inside and not let it escape! So, the camera wouldn't come back.

Also, the camera would stop working as it gets closer to the black hole because the black hole's super-strong pull (gravity) would break it. Even if the camera somehow kept working and came back, the pictures or videos it took would be all strange and stretched, not like anything we see around us. This is because black holes bend light and time in weird ways.

Right now, this idea is more like a fun science fiction story because we don't have the technology to do it, and black holes are really, really far away from us."\\
\midrule
\textbf{GPTZero result}: AI

\textbf{GPTZero explanations}: Readability: 72.3(High) Percent SAT: 1.7(Medium) Simplicity: 35.2(low) Perplexity: 45.3(Medium) Burstiness: 37.9(Medium) Average sentence length: 22.3(Medium)\\
\bottomrule
\end{tabular}
\caption{Comparison between abstract scores from GPTZero and human-readable explanations}
\label{tab:tablef2}
\end{table*}

\section{Examples of ``Undecided'' and Human Annotators' Explanations}
\label{appendix:binary_vs_ternary_more_examples}

Due to space limitations in the main text, we provide additional examples here to illustrate texts categorized as ``undecided'' along with the explanation notes given by human annotators. These examples demonstrate cases where human readers identified stylistic or structural signals consistent with either authorship type, but lacked decisive features pointing strongly toward human or machine generation.

\begin{table*}[h]
\centering
\small
\begin{tabular}{@{}p{\linewidth}@{}}
\toprule
\textbf{Text}: Listen, I've been in your shoes before, and the best advice I can give you is to embrace change. Life is unpredictable, and sometimes we get comfortable in our routines, but growth happens when we step out of our comfort zones. Don't be afraid to take on new challenges, explore different opportunities, and learn from every experience, even if it seems daunting at first. Remember, the magic happens outside your comfort zone. So, be open to change, embrace the unknown, and trust in your ability to adapt. You'll be amazed at the personal and professional development that follows.\\
\midrule
\textbf{Human explanation}: The language is clear, smooth, and well-structured, with no surface-level mistakes. The tone is encouraging and empathetic, employing rhetorical strategies like direct address and motivational phrasing. These features suggest human authorship, but the overall clarity and polish could also reflect advanced model output. Because neither the personal voice nor the linguistic imperfections are strong enough to determine origin, the annotator considered it ``undecided''.\\
\bottomrule
\end{tabular}
\caption{An ``undecided'' text where both interpersonal tone and formal consistency coexist.}
\label{tab:tableg1}
\end{table*}

\begin{table*}[h]
\centering
\small
\begin{tabular}{@{}p{\linewidth}@{}}
\toprule
\textbf{Text}: To conclude, we empirically show that a significant number of later layers of CNNs are robust to the absence of the spatial information, which is commonly assumed to be important for object recognition tasks. Modern CNNs are able to tolerate the loss of spatial information from the last 30\% of layers at around 1\% accuracy drop; and the test accuracy only decreases by less than 7\% when spatial information is removed from the last half of layers on CIFAR100 and Small-ImageNet-32x32. Though depth of the network is essential for good performance, the later layers do not necessarily have to be convolutions.\\
\midrule
\textbf{Human explanation}: The writing is technically accurate and domain-specific, using precise numerical details and scientific phrasing typical of expert authorship. The logical flow and factual completeness indicate deep knowledge. However, the dense and impersonal tone, along with consistent structure and polished syntax, resembles model-generated summaries. Given the lack of strongly distinctive human voice or error, the annotator classified it as ``undecided''.\\
\bottomrule
\end{tabular}
\caption{An ``undecided'' text where technical correctness and neutral tone lead to ambiguity.}
\label{tab:tableg2}
\end{table*}

\section{Quantitative Representation of Explanations from Human Annotators}
\label{appendix:quantitative_human_explanations}

%Table~\ref{tab:tableh1} presents the average perplexity and readability scores for classes labeled by different human coders. It is evident that texts classified as human-written exhibit higher perplexity and readability scores, indicating a more complex and readable nature in comparison to texts labeled as undecided or AI-generated.

We quantified the perplexity and readability of explanations provided by human coders. Text perplexity was computed using scripts from the Natural Language Toolkit (NLTK) with the GPT-2 model, while readability was measured using the Flesch Reading Ease and Flesch-Kincaid Grade Level formulas.

%The distributions of perplexity, Flesch Reading Ease, and Flesch-Kincaid Grade Level for different categories classified by human coders are illustrated in Fig.~\ref{fig:figh1}, Fig.~\ref{fig:figh2}, and Fig.~\ref{fig:figh3}, respectively. 

Table~\ref{tab:tableh1} presents the average values of these measures for texts in each category. It is evident that texts classified as ``AI'' by human coders exhibit lower perplexity and lower readability scores. Specifically, a higher Flesch-Kincaid Grade Level value indicates a higher required English proficiency level, which corresponds to a lower readability score. Texts classified as ``undecided'' fall between the ``human'' and ``AI'' categories. Thus, our annotations by human coders are validated.

\begin{table*}[tb]
\centering
\small
\begin{tabular}{lccc}
\toprule
Category & Perplexity & Flesch Reading Ease & Flesch-Kincaid Grade Level\\
\midrule
Human     & 52.72 & 69.42 & 7.95\\
Undecided & 34.21 & 57.44 & 9.28\\
AI        & 21.62 & 48.02 & 10.72\\
\bottomrule
\end{tabular}
\caption{Average perplexity and readability scores for different classes labeled by human coders. A higher Flesch Reading Ease score indicates greater readability, while a higher Flesch-Kincaid Grade Level score indicates lower readability.}
\label{tab:tableh1}
\end{table*}

\section{More Details about Human Annotators' Work}
\label{appendix:human_annotation}

The three human annotators we used are all co-authors of the work. We did not recruit any other human participants for the annotation task because the construction of the ternary dataset required two rounds of iteration, including a second round of discussions among all authors to help the three annotators reach a consensus. We considered this approach more appropriate for our study than using external annotators. Since the annotation work was conducted by co-authors and did not involve external participants, the study did not require approval from our institution’s research ethics review board. The annotators were not financially compensated, as their contribution was considered part of their technical involvement in the research.

To minimize bias in the annotation results, each annotator first completed the task independently. Afterward, they convened to resolve any disagreements through discussion. For the independent annotation phase, we provided each annotator with the unlabeled dataset and detailed instructions, as shown in Table~\ref{tab:annotation_instructions}. During the discussion phase, the first author facilitated the process and consulted with other co-authors regarding borderline cases to support fair and consistent final decisions.

\begin{table*}[h]
\centering
\small
\begin{tabular}{@{}p{\linewidth}@{}}
\toprule
The file in the experiment folder is a spreadsheet where we record the text, detection results, and explanation of detection results.

Your task is to determine the likely source of the text in the first column and provide an explanation for your judgment.

The file contains three columns:

The first column is ``text,'' which contains the text to be analyzed. Each piece of text should be categorized as one of the following: ``human,'' ``machine,'' or ``undecided.''

The second column is ``detection result,'' where you indicate your judgment on the source of the text. Label the text as ``human'' if you believe it was written by a person, ``machine'' if you believe it was generated by a model, and ``undecided'' if the source cannot be clearly determined.

The third column is ``explanation,'' where you briefly explain your reasoning behind the label in ``detection result.''

Below is an example for illustration:

\textbf{Text}: A fan is an electrical appliance used for cooling and air circulation. It operates by rotating blades, which create a breeze to cool down a room or space. Fans come in various types, including ceiling fans, table fans, and pedestal fans, each designed for specific needs. They are energy-efficient and provide a cost-effective way to stay cool, especially during hot weather. Fans also help in ventilating areas by moving stale air and introducing fresh air.

\textbf{Detection result}: undecided

\textbf{Explanation}: This text discusses the topic of electric fans in a neutral, factual manner. The structure is clear and free from grammatical or logical issues. It shows characteristics common to both human- and machine-generated writing, making it difficult to confidently determine the source. Therefore, I marked it as ``undecided.''

\textbf{Note}: The labeled results are intended solely for academic research.\\
\bottomrule
\end{tabular}
\caption{Human Annotation Instructions}
\label{tab:annotation_instructions}
\end{table*}

\end{document}